\documentclass[sn-mathphys,Numbered]{sn-jnl}% Math and Physical Sciences Reference Style
\usepackage{graphicx}%
\usepackage{multirow}%
\usepackage{amsmath,amssymb,amsfonts}%
\usepackage{amsthm}%
\usepackage{mathrsfs}%
\usepackage{multicol}
\usepackage[title]{appendix}%
\usepackage{xcolor}%
\usepackage{textcomp}%
\usepackage{manyfoot}%
\usepackage{booktabs}%
\usepackage{algorithm}%
\usepackage{algorithmicx}%
\usepackage{algpseudocode}%
\usepackage{listings}%
\usepackage{comment}
\RequirePackage{arabtex}
\RequirePackage[utf8]{inputenc}
\RequirePackage{utf8}
\setcode{utf8}
\usepackage{adjustbox}
\usepackage{makecell}
\usepackage{multirow}
\usepackage{hhline}
\usepackage{array}
\newcolumntype{H}{>{\setbox0=\hbox\bgroup}c<{\egroup}@{}}
\theoremstyle{thmstyleone}%
\theoremstyle{thmstyletwo}%

\theoremstyle{thmstylethree}%

\begin{document}

\title[Arabic Legal Judgment Prediction]{ALJP: An Arabic Legal Judgment Prediction in Personal Status Cases Using Machine Learning Models}

\author*[1]{\fnm{Salwa} \sur{Abbara}}\email{sabbara@stu.kau.edu.sa}
\equalcont{These authors contributed equally to this work.}

\author[1]{\fnm{Mona} \sur{Hafez}}\email{mhafez0002@stu.kau.edu.sa}
\equalcont{These authors contributed equally to this work.}

\author[1]{\fnm{Aya} \sur{Kazzaz}}\email{akazzaz@stu.kau.edu.sa}
\equalcont{These authors contributed equally to this work.}

\author*[1]{\fnm{Areej} \sur{Alhothali}}\email{aalhothali@kau.edu.sa}
\equalcont{These authors contributed equally to this work.}
\author[2]{\fnm{Alhanouf} \sur{Alsolami}}\email{aamalsolami@kau.edu.sa}
\equalcont{These authors contributed equally to this work.}

\affil*[1]{\orgdiv{Computer Science Department}, \orgname{Faculty of Computing and Information Technology, King Abdulaziz University}, \orgaddress{\street{Street}, \city{Jeddah}, \postcode{21589}, \country{Saudi Arabia}}}

\affil*[2]{\orgdiv{Faculty of Law}, \orgname{King Abdulaziz University}, \orgaddress{\street{Street}, \city{Jeddah}, \postcode{21589}, \country{Saudi Arabia}}}

\abstract{Legal Judgment Prediction (LJP) aims to predict judgment outcomes based on case description. Several researchers have developed techniques to assist potential clients by predicting the outcome in the legal profession. However, none of the proposed techniques were implemented in Arabic, and only a few attempts were implemented in English, Chinese, and Hindi. In this paper, we develop a system that utilizes deep learning (DL) and natural language processing (NLP) techniques to predict the judgment outcome from Arabic case scripts, especially in cases of custody and annulment of marriage. This system will assist judges and attorneys in improving their work and time efficiency while reducing sentencing disparity. In addition, it will help litigants, lawyers, and law students analyze the probable outcomes of any given case before trial. We use a different machine and deep learning models such as Support Vector Machine (SVM), Logistic regression (LR), Long Short Term Memory (LSTM), and Bidirectional Long Short-Term Memory (BiLSTM) using representation techniques such as TF-IDF and word2vec on the developed dataset. Experimental results demonstrate that compared with the five baseline methods, the SVM model with word2vec and LR with TF-IDF achieve the highest accuracy of 88\% and 78\% in predicting the judgment on custody cases and annulment of marriage, respectively. Furthermore, the LR and SVM with word2vec and BiLSTM model with TF-IDF achieved the highest accuracy of 88\% and 69\% in predicting the probability of outcomes on custody cases and annulment of marriage, respectively.}




\maketitle

\section{Introduction}
\label{sec:introduction}

%\lipsum[2-4]~\citepp{zhang2016exploring,Aniket2016,7178180,8551517,cocsar2016toward,piciarelli2008trajectory}.
With the development of artificial intelligence (AI) and judicial informatization reform, the development of intelligent justice has received a great deal of attention. The proper application of artificial intelligence technology can improve judicial practitioner efficiency, optimize justice methods, and reduce sentence inconsistency. Several intelligent solutions, such as intelligent legal document generation, automatic question answering, and automatic speech recognition in the court system, have been effectively implemented in casework. Legal judgment prediction (LJP) is the critical point of artificial judicial intelligence. LJP systems aim to predict judgment results according to the facts of cases with feasible judgment suggestions, such as the prediction of charges, imprisonment terms, and applicable law articles. LJP systems can also assist litigants, attorneys, students, and teachers by improving their work and time efficiency while reducing the risk of making mistakes, so it serves as a valuable resource for professionals.

Being in the legal field is a boatload of predicting and the risk of facing various hidden dangers while doing so. When deciding whether or not to take a new case, an astute attorney would evaluate the nature of the legal issues at hand and the case's likely results. Typically, wise attorneys would avoid making precise, unambiguous predictions about what they likely believe the conclusion will be. They safely distanced themselves from a seemingly ironclad prediction, emphasizing that they do not guarantee the legal results. New attorneys often make bold predictions; they almost end up on the wrong side of a displeased client who later remembers how confidently boastful the first prediction was, especially if the case goes astray from the anticipated outcome. The advent of Artificial Intelligence (AI) in the legal industry enhances attorneys' prediction abilities. It becomes more practical to analyze a huge corpus of legal cases and generate predictions for a newly given legal case utilizing AI techniques, such as Natural Language Processing and Machine Learning. Using AI and deep learning in LJP is beneficial in providing fast decision-making and outcomes which allows input data to be instantly verified, unbiased judgment towards gender or nationality, analysis with previous instances of comparable patterns, and identify situations with substantial variation in human and AI choices makes it easier to uncover corruption. 

Several studies have investigated the possibility of predicting judgment outcomes in legal cases written in English, Hindi, and Chinese. No study has investigated the prediction of legal judgment in Arabic,  especially in personal status cases such as custody, divorce, and annulment of marriages. Thus, we develop an LJP system that performs two tasks in this paper. The first task is to predict judgment decisions and relevant law articles or evidence in (custody and annulment of marriage) using the pleading of the case. The second is to predict the probability of possible outcomes of personal status (custody and annulment of marriage) cases given the plaintiff's claim and defend answer. We developed an Arabic legal prediction dataset for personal status cases (using the Kingdom of Saudi Arabia (KSA) as a case study). The dataset was generated using a sample collection published Ministry of Justice and a simulated dataset by experts in the domain. We evaluated the proposed model using different machines and deep learning models such as SVM, LR, LSTM, and BiLSTM with different data representation techniques such as TF-IDF and word2vec on the developed dataset. %the dataset published by the Saudi Ministry of Justice and the data that we simulated.

\section{Related work}\label{sec:Related}

%\subsection{Machine Learning Models}
Several studies have utilized machine learning techniques to predict judicial outcomes. \citet{aletras2016predicting} presented the first comprehensive research of predicting the result of cases heard by the European Court of Human Rights (ECHR) based only on the textual content presented in the court to determine if a human rights article has been violated or not. An Support Vector Machine (SVM) classifier trained on textual information using N-grams and topics in Human Rights Documentation cases was used, achieving high accuracy in predicting court outcomes (79\%). \citet{sil2020novel} also proposed a model which aims to deliver justice by providing judicial argument-based analysis using the SVM algorithm. The model is trained on  features like years of marriage, dowry details, and postmortem reports from a dataset of ‘dowry death’ cases in West Bengal. The model achieved 93\% accuracy in binary classification.  \citet{medvedeva2020using} proposed a machine learning model to predict court decisions using textual information from ECHR cases. The program uses SVM Linear classifier and word n-gram TF-IDF to analyze textual data, predicting 75\% of cases correctly. 
%proposed a model to predict the court's decision by extracting textual information from relevant sections of the ECHR judgments and estimate the potential of predicting future cases. For this reason, they developed a computer program that analyzes texts of ECHR judgments published on the court's website and predicts if any specific article of the ECHR was violated. For this task, they use SVM Linear classifier. They used a word n-gram and Term Frequency - Inverse Document Frequency (TF-IDF) to represent textual data. As a result, 75\% of the cases were predicted correctly. 
Similarly, \citet{shaikh2020predicting} proposed a machine learning model to predict outcomes of murder-related cases in the Delhi  District  Court. Several machine learning classifiers were evaluated in this task, including Classification and Regression Trees (CART), Bagging, Random Forest, and SVM, to predict trial outcomes using several handcrafted features obtained by manually analyzing the cases. The results show that the best performance is obtained using Bagging, RF, and SVM, CART with ranges between 91,86\% and 90.70\%. %The findings indicate the importance of the factors studied and their ability to predict outcomes. However, this appears to be a weakness though i.e. poor performance in predicting outcomes in cases when there are several defendants and the outcome is different for each.

%%%%%%%%%%%%%%%%%%%%%%%%%%%%%%%%%%%%%%%%%%%%%%%%%%%%%%%%%%%%%%%%%%%%%%%%%%%%%%%%%%%%%%
%\subsection{Deep Learning Based Models}

Several studies have utilized deep learning techniques in predicting legal case outcomes. \citet{Zhang2019ApplyingDD} proposed an automatic law article prediction model based on Deep Pyramid Convolutional Neural Network. They predict the relevant law article for cases using case description and associated legal provisions. %They constructed the data discretization pattern and replaced the original numerical content with interval labels, allowing the model to recognize the specific meaning of numerical data of various sizes. The processed data is then fed into a \gls{DPCNN} model that can successfully detect long-distance text dependencies. 
The results show that the proposed method outperforms various state-of-the-art baselines models on several public datasets~\cite{xiao2018cail2018}.  \citet{li2019element} %proposed element-aware LJP approach, which aims to extract the corresponding legal constitutive elements as discriminative features to distinguish confusing charges. They 
proposed a neural network model based on an element-driven attention mechanism that takes the textual description of a criminal case as the input and predicts the charges, applicable law articles, and prison terms. The approach is evaluated on a real-world dataset containing $125,830$ judgment documents of criminal cases published by China Judgments. % They randomly selected 70\% of all the cases for training, 15\% for validation, and 15\% for testing. 
%The  results showed that the proposed model could be effectively applied to criminal cases with confusing charges. 
The model accuracy for element prediction, charge prediction result, law article prediction, and prison term was 98.83\%, 97.92\%, 98.16\%, and 82.13\%, respectively. Most existing methods follow the text classification framework that fails to model the complex interactions among complementary case materials.  \citet{long2019automatic} formalized the task as Legal Reading Comprehension according to the legal scenario. The framework predicts the final judgment results based on three types of information, including fact description, plaintiffs’ pleas, and law articles and predict if a certain plea in a given civil case would be supported or rejected. %To instantiate this framework, they proposed an end-to-end neural network model. They randomly collected 100,000 cases from China Judgments Online, among which 80 000 cases are for training 10,000 each for validation and testing. They did data preprocessing such as (Name Replacement, Law Article Filtration). Experimental results show that the model achieves considerable improvement over all the baselines and the accuracy of this model gives 82.2\% .

%%%%%%%%%%%%%%%%%%%%%%%%%%%%%%%%%%%%%%%%%%%%%%%%%%%%%%%%%%%%%%%%%%%%%%%%%%%%%%%%%%%%%%

%\subsection{Multi-task Learning Based Models}

Multi-task learning (MTL) models were also examined for LJP in several studies. The MTL model considers the relationship between subtasks such as law articles, charges, and penalty terms in LJP task. \citet{zhong2018legal} proposed a topological multi-task learning framework that formulates the dependencies among subtasks (law articles, charges, fines, and the term of penalty) as directed acyclic graph (DAG) to jointly predict the trial subtasks outcomes. Their approach outperforms single-task baselines and conventional multi-task learning models on three Chinese criminal case datasets. \citet{wang2020study} proposed a MTL LJP model based on CNN-BiGRU and is used to improve the accuracy and efficiency of legal judgment prediction. Several data representation were used including TF-IDF and word2vec, word embedding, fact encoding, and document representation achieving accuracy of $95.1\%$ for Law articles prediction, $95.2\%$ for Charges, and $72.6\%$ for the term of penalty.  %This model improves the prediction performance of \gls{CNN} or \gls{RNN} alone 
\citet{huang2019model} built a multi-task deep neural network classification model based on integrating CNN with the attention LSTM model to achieve high precision of crime and related law prediction. %This study took the criminal case as input, and the main goal was to make the model accurately predict the charge and relevant law articles involved in the case.
The model is evaluated on a dataset of $626,600$ judicial documents collected from the Internet achieving an average F1 score of 93.62\% and 90.84\% for crime, related law prediction, respectively. 

\citet{li2019mann} proposed a multi-channel attentive neural network that uses attention mechanism and BiGRU hierarchical sequence encoder to learn better semantic representation and interaction among different parts of case descriptions. %The model takes takes the fact description and defendant persona as input and generates two candidate sets of relevant law articles. 
The multi-channel attentive encoders take a generated law articles based on the case fact description and defendant persona as input and pass them into three hierarchical encoders (fact-channel encoder, persona channel encoder, article-channel encoder) incorporating with word-level and sentence-level attention context vector to predict the charges and prison term. %The approach evaluated on four datasets . %They construct four dataset named as \gls{CJO-S}, \gls{CJO-L}, and \gls{PKU}. The F1 score for the proposed framework \gls{MANN} on the \gls{RCAIL} dataset is 91.2\% for charges, 82.5\% for law articles, and 40.2\% for a prison term. Also, the F1 score on the \gls{CJO-S} dataset is 90.1\% for charges, 84.2\% for law articles, and 41.2\% for a prison term. And the F1 score on the \gls{CJO-L} dataset is 92.3\% for charges, 87.2\% for law articles, and 42.9\% for a prison term. Finally, on the \gls{PKU} dataset the F1 score is 88.7\% for charges, 82.9\% for law articles, and 40.9\% for prison term.\\
\citet{yao2020gated} proposed a novel gated hierarchical multitask learning network to jointly model multiple sub-tasks (law article, charge, and term of penalty) in judicial decision prediction. The model combines a Gated Hierarchical Encoder (GHE) to extract in-depth semantic information of fact description from multiple perspectives and a Dependencies Auto-learning Predictor (DAP) to learn the dependencies among sub-tasks dynamically. The proposed model takes the fact description as input and predicts law article, charge, and term of penalty. %The authors used the CAIL2018-Large and CAIL2018-Small datasets~\cite{xiao2018cail2018} for evaluation. 

%%%%%%%%%%%%%%%%%%%%%%%%%%%%%%%%%%%%%%%%%%%%%%%%%%%%%%%%%%%%%%%%%%%%%%%%%%%%%%%%%%%%%%%%

%\subsection{Attention Mechanism Based Model}
The attention mechanism has been successfully used in many NLP tasks in recent years. A number of studies used attention mechanisms in the task of predicting judicial outcomes. \citet{bao2019charge} proposed an attention neural network that uses relevant articles to improve the performance and interpretability of charge prediction tasks. The model uses the fact description to extract relevant law articles that assist in locating key information from the fact description and help improve the performance of charge prediction. % The data was collected from the first large-scale Chinese legal dataset CAIL 2018.% They randomly chose 203,823 cases for training, 20,000 for validation, and 40,000 for testing. %In addition, to model the multi-label property in real-world scenarios, they kept data with multiple charges or relevant articles, which accounted for 18.6\%, 10.5\%, and 16.7\% of the training set, validation set, and test set, respectively. Micro-F1 for relevant article ex-traction was 78.7\% and 81\% for charge prediction. 
%The attention mechanism has been successfully used in many NLP tasks in recent years. A number of studies used attention mechanisms in the task of predicting judicial outcomes.
 To address the challenges of predicting judgment in lengthy cases, \citet{Sukanya2021} presented an effective hierarchical attention deep neural network model with fine-tuned transformer to predict legal cases outcomes. %They review the challenges faced by the judgment prediction system due to lengthy case facts using the deep learning model. Case Facts of the real World have two main challenges. One is the difficulty faced in encoding lengthy documents, and the other is the lack of full external information. Existing models used for text classification and prediction like \gls{RNN} and \gls{LSTM} work sequentially and are predicted only based on case facts without any consideration of any external information such as evidence and emotions. In addition, they are applicable for a single defendant person. Hierarchical Attention Neural network models with fine-tuned transformer concepts will give an efficient improvement based on quality and time in judgment prediction. Hierarchical attention models are used mainly to automatically read and extract information from case facts efficiently. On the other side, the main benefit of using transformer-based models is: that the input tokens are not processed sequentially one by one. Also, labeled data is not necessary. The main benefit of Hierarchical Attention Neural network models with fine-tuned transformers is that they can cope with any application using its good contextualized inbuilt word embedding, which supports large words \citet{Sukanya2021}. 
 \citet{yang2019recurrent} employed LSTM with self-attention to simulate a judge's recurrent reading behavior utilizing semantic mutual information between evidence and article. %\gls{RAN} takes care of recurrent semantic interactions between fact descriptions and articles. The RAN employs \gls{LSTM} and self-attention to embed both articles and facts into a small embedding space. The main difference between \gls{RAN} and other attention mechanisms is that they employed the article definition as external information and used a recurrent attention block to capture multiple repeated interaction attention information between fact and article to support judgment prediction. The encoder, recurrent, and output layers are the three layers that form the \gls{RAN}. They use shallow, neural network-based, and attention-based models such as \gls{KNN}, \gls{BR}, \gls{CC}, \gls{CNN}, \gls{BiLSTM}, and \gls{DPAM} as baselines for comparison and evaluation of the \gls{RAN} model performance. Jaccard similarity coefficients and Macro-averaging (macro-precision, macro-recall, macro-F1) are used to evaluate the performance of models. Experiments on real-world datasets \gls{CJO}, \gls{CAIL} small, and \gls{CAIL}2018 show that the RAN model outperforms state-of-the-art methods significantly\citep*{yang2019recurrent}.
 %Similarly, \citet{bao2019charge} proposed an attentional neural network which uses relevant articles to improve the performance and interpretability of charge prediction tasks. The automatic charge prediction task is to use the fact description to extract relevant articles; In return, the selected relevant articles assist in locating key information from the fact description, which helps improve the performance of charge prediction. 
\citet{xu2020distinguish} presented an end-to-end model to solve the task of LJP. To distinguish confusing charges, they propose a novel graph neural network to automatically learn subtle differences between confusing law articles by capturing essential but rare features and design a novel attention mechanism that fully exploits the learned differences to extract compelling discriminative features from fact descriptions attentively.% They used the publicly available datasets \gls{CAIL}2018, \gls{CAIL-S}, \gls{CAIL-B}. First, use law distillation model to divide law articles into several communities, which consist of \gls{GCL} and \gls{GDO} to remove duplicated texts between two law articles and only use the leftover texts for the attention mechanism. Then re-encoding Fact with  Distinguishable attention. Finally, using a linear classifier for law article prediction, charge prediction, and term of penalty prediction. They use their method LADAN with the same multi-task framework (\gls{LADAN} + \gls{MTL}, \gls{LADAN} + TOPJUDGE, \gls{LADAN} + \gls{MPBFN}). The accuracy for the proposed framework \gls{LADAN} on the \gls{CAIL-S} dataset with \gls{MTL} is 81.2\% for Law articles, 85.07\% for Charges, and 38.29\% for the term of penalty. Also, the accuracy of \gls{LADAN} + TOPJUDGE on the same dataset is 81.53\% for Law articles, 85.12\% for Charges, and 38.34\% for the term of penalty. Also, the accuracy of \gls{LADAN} + \gls{MPBFN} on the same dataset is 82.34\% for Law articles, 84.83\% for Charges, and 35.39\% for the term of penalty. The accuracy on the \gls{CAIL-B} dataset with \gls{MTL} is 96.57\% for Law articles, 96.45\% for Charges, and 59.66\% for the term of penalty. Also, the accuracy of \gls{LADAN} + TOPJUDGE on the same dataset is 96.62\% for Law articles, 96.39\% for Charges, and 59.70\% for the term of penalty. Also, the accuracy of \gls{LADAN} +\gls{MPBFN} on the same dataset is 96.6\% for Law articles , 96.42\% for Charges, and 59.85\% for term of penalty~\citep*{xu2020distinguish}.

\citet{kowsrihawat2018predicting} proposed a prediction model of criminal cases using End-to-End BiGRU deep learning neural networks. Their model imitates a process of legal interpretation, whereby recurrent neural networks read the fact from an input case and compare them against relevant legal provisions with the attention mechanism. %The model’s output shows if a person is guilty of a crime according to the fact and laws. They innovate the open dataset called \gls{TSCC}. Their expert read and extracted some information from \gls{TSCC}. The dataset consists of two data tables called \gls{TSCC} Judgement (1207 records)and \gls{TSCC} Law (122 records). The model takes the Fact text and Law text (text of relevant legal provision) as inputs to predict a judicial decision. The output has a binary value, one means guilty, and zero means innocent. As a result, this model achieves 63.35\% as Macro F1 and 66.67\% as Micro-F1 score for both classes. The dataset was unbalanced because the number of not-guilty in the dataset was greater than that of guilty, which may affect the accuracy of the proposed model.\\
Some related studies have combined several deep neural models. ~\citet{yuan2019automatic} proposed a framework for the automated judging based on ensemble strategy that combines many  deep neural network models and manual law features to solve the problem of data imbalanced. Simultaneously, they increase the framework's performance by enhancing the data. Their model shows that data enhancement and ensemble strategy can improve the performance of judgment prediction.
%%%%%%%%%%%%%%%%%%%%%%%%%%%%%%%%%%%%%%%%%%%%%%%%%%%%%%%%%%%%%%%%%%%%%%%%%%%%%%%%%%%%%%%%
%\subsection{Transformer-based Models}

Transformer-based models have shown a tremendous impact on many NLP problems. Similar to other NLP problems, transformer-based models were successfully used in predicting the outcome of the legal cases. \citet{wang2020deep} utilized the recently widely used pre-trained language model Bidirectional Encoder Representations from Transformers (BERT) for LJP. %They use BERT and comapred  with other deep learning models such as CNN, LSTM, DPCNN, and RCNN to predict accusations in each case. %The dataset comes from criminal legal documents published by China Judgment Online, with 219.608 as the size of the dataset. They divide the dataset into a training set, testing set, and validation set. The training set includes 158,117 data, the validation set includes 17,569 data, and the test set includes 43,922 data. Each piece of data consists of the description of the case and the facts in the legal document, as well as the legal provisions involved in each case, the accusations that the defendant was convicted of, and the length of the sentence. They did data preprocessing such as (segmentation, deleting stop words, and extracting keywords), then they used a pretraining language representation model which produced the embedding vector, and they fed this vector to multiple deep learning models. 
BERT model significantly improves accusation prediction accuracy compared to other   deep learning models with Word2Vec representation. %The recall and the F1 for BERT+FC were 89.79\% and 89.64\%, respectively~\citep*{wang2020deep}.
Similarly, \citet{chalkidis2019neural} developed a hierarchical version of BERT for judgment prediction. They present a new publicly available dataset of English legal judgment prediction cases from the ECHR. %They evaluated multiple neural models such as \gls{BiGRU-Att}, \gls{HAN}, \gls{LWAN}, \gls{BERT} and \gls{HIER-BERT} on the new dataset. Three tasks were considered when classifying the data. The first task is binary classification to determine whether human rights have been violated or not. The second task is case importance detection. While the third task is a multi-label classification to determine the type of violation if any. They utilized a data anonymization technique to investigate whether legal prediction models are biased towards demographic data or factual information important to human rights. Micro-averaged precision (P), recall (R), and F1 were used for evaluation. Neural models surpass \gls{SVM}s in all tasks with bag-of-words, the only method tested in English legal judgment prediction so far. The micro-averaged precision, recall, and F1 for the hierarchical version of \gls{BERT} were 65.9\%, 55.1\%, and 60\%, respectively. Although the neural models are considered to outperform previous feature-based models but provide no justification for their predictions plus in labeling with FEW training examples. All models underperform, and illustrate the difficulties of few-shot learning in \gls{ECHR} legal judgment prediction \citep*{chalkidis2019neural}.\\  
\citet{zhu2020legal} proposed a Transformer-Hierarchical-Attention Multi-Extra Network that takes fact description, court view, and basic information of the defendant as input to predict law articles, charges, and terms of penalty. The dataset they used is CJO consists of criminal cases published by the Chinese government from China Judgment Online. 

%_______________________

%-------------------------------------------
%\subsection{Tensor Decomposition Based Model}
Some studies have employed Tensor decomposition techniques in legal cases outcome prediction. \citet{guo2020tenrr} developed a new algorithm based on innovative tensor decomposition and ridge regression for judgment prediction. The model tested on a dataset obtained from the Chinese Referee Document Network.% It contains nearly 3 million legal cases involving 203 crimes%. The first module is \gld{RTenr} which represents each legal case as a three-dimensional original tensor and manually extracts case features. The second module is \gls{ITend} which decomposes the original tensor representing a legal case into a core tensor by removing substantial inaccurate, meaningless, and redundant information using mapping matrices to avoid data sparseness and dimensional explosion. And the third module is \gls{ORidge} which controls the tensor decomposition process in ITend through mapping matrices. The obtained core tensors carry tensor elements and tensor structure information that is most conducive to improving the accuracy of TenRR. The dataset used was obtained from the Chinese Referee Document Network. It contains nearly 3 million legal cases involving 203 crimes. As an input, the model takes the legal case and predicts the judgment as sentences and fines. The accuracy of the predicted Sentences in legal cases of fixed-term imprisonment is 93.71\%, life imprisonment is 91.62\%, the death penalty is 92.6\%, and fines in legal cases is 93.56\% .
In the same vein, \citet{guo2019rnrtd} build an intelligent judgment approach, which is based on the relationship-driven recurrent neural network and restricted tensor decomposition. %They present legal case data as tensors and then use \gls{RTD} technique to extract valuable tensor features from legal case tensors and reduce tensor dimensionality. 
The recurrent neural network were used for intelligent judgment of multiple accusations in legal cases. The model tested on legal cases obtained from a Chinese refereeing study network.% \gls{rdRNN} controls the impact of cases similar to the current case on the output status of the current case. They use nearly 1.8 million historical legal cases obtained from a Chinese refereeing study network. They use \gls{BiLSTM} as a basic algorithm with \gls{RnRTD}, and the accuracy was 93\%, but algorithms based on \gls{RNN}s run very slowly, and the computational complexity of it needs to be reduced.\\

Previous studies in LJP have utilized various machine and deep learning techniques, including SVM, LR, and K-Nearest Neighbors. Most studies utilized deep learning techniques and sequence models, while some used attention mechanisms for predicting legal case outcomes. Tensor decomposition-based models were also employed. LJP models rely on law articles and predict law articles, charges, and penalties. Researchers have formulated problems as single-task learning or multi-task learning, predicting single or multiple outcomes. Previous studies were mainly conducted in Chinese, Hindi, and English, and mostly focused on crime cases. However, no studies have been conducted in Arabic or personal status cases, as personal status cases in KSA are adjudicated at the judge's discretion, not on law articles.

\section{Dataset}

Several dataset were used in the filed of LJP. These dataset were generally written in Chinese, English, and Hindi. To serve the purpose of this study, we developed an Arabic LJP dataset for Saudi Arabian Personal Status Cases, we have first collected personal status cases published by the ministry of justice and we also generated a new sample of cases throughout experts in the field. Table~\ref{Ministry of Justice dataset} shows information about the Ministry of Justice dataset. Each article in the sample collection  contains the topic which provide details of case category, the evidence which provide details of the reasons and legal evidence that was relied upon in judging the case, a  summary of the case which include a summary of the claim, answer, and judgment, the final judgment which is a description of the claim, answer, and sessions that took place in the court, in addition to the judgment on the case and the. The sample collection has five type of cases which are custody, annulment of marriage, visiting, divorce, and alimony. We have only consider in this research custody and annulment of marriage cases due to limitation and variation of outcomes of the rest of the personal status cases in the current sample of collection. The dataset has a total number of $49$ cases that each has a true judgment outcome. 

The dataset textual content were obtained from portable document format(PDF) files and many words are extracted incorrectly due to the font in which the data was written.  Thus, to correct some of the extracted data, several data cleaning and manual corrections were performed. %Sample of the corrections are shown in Table~\ref{Wrong vs correct words}.  

\begin{table}[h]
\begin{tabular}{|c|c|}
%\hline
% \multicolumn{2}{|c|}{Available Dataset}\\
 \hline
 \textbf{Source}&Ministry of justice\\
 \hline
 \textbf{Content}& Topic, evidence, case summery, judgment\\
 \hline
 \textbf{Types} & \textbf{Numbers of Cases}\\ 
 \hline
 Custody*&20\\
 \hline
 Annulment of marriage*&29\\
 \hline
 Visiting&2\\
 \hline
 Divorce&0\\
 \hline
 Alimony&44\\
 \hline
\end{tabular}
\caption{Ministry of Justice Sample collections}
\label{Ministry of Justice dataset}
\end{table}

%\begin{table}
%\centering
%\begin{tabular}{|c|c|}
%\hline
%wrong word & corrected word \\
%\hline

%\RL{املدعية}
%&
%\RL{المدعية}
%\\
%\hline

%\<املدعى>
%&
%\<المدعى >\\
%\hline

%\<املولودة >
%&
%\<المولودة >\\
%\hline

%\end{tabular}
%\caption{Wrong vs correct words}
%\label{Wrong vs correct words}
%\end{table}

We have formulated the problem as a multi class classification problem. The multi class classification task, takes a pleading text as input and generate the judgment (one of multiple outcomes) and reasons/evidences (one of multiple reasons) as output. The original trail cases were analyzed with experts in the law field (lawyers and professors) and a list of possible and more common judgement decisions were obtained. Based on this analysis, we formulated the problem of predicting judgement in custody cases as a multi-class classification problem of four classes, namely, mother grant full custody, father grant full custody, children over seven years old have parental choice, while those under seven have custody with their mother until they reach seven, and other. Table~\ref{Classes in custody Cases} shows the judgment classes in custody cases.  For the reasons or law article prediction, we formulated the problem as a multi class classification with eight classes correspond to the law articles used in the custody cases. For annulment of marriage cases, we had also four classes which are annulment of marriage with compensation, annulment of marriage without compensation, deny annulment, other. Table~\ref{Classes in annulment of marriage Cases} show the classes of annulment of marriage. The reasons or law article prediction in the annulment of marriage cases were formulated as a multi class classification problem with 11 classes.

Table~\ref{Data frame for single label classification} shows an example of the binary classification problem.
\renewcommand{\tabcolsep}{2pt}

  \begin{table}
                \centering
                \begin{tabular}{|c|c|}
                \hline
                Class Name in Arabic &Class Name in English\\
                \hline
                \RL{تخيير الابناء فوق السبع سنوات، وتكون الحضانة للام لمن لم يبلغ سبعه سنوات} & \thead{Children over seven years old have parental choice,\\ while those under seven have custody with their\\ mother until they reach seven}
\\
                 \hline
                \RL{حضانة الاولاد لوالدتهم}& mother grant custody of children\\
                \hline
                \RL{حضانة الاولاد لوالدهم}& mother grant custody of children\\
                \hline
                \RL{أخرى} & Other\\
                \hline
                \end{tabular}
                \caption{Judgement Decision Classes in Custody Cases}
                \label{Classes in custody Cases}
                \end{table}

\begin{table}
  \begin{tabular}{|c|c|}
                \hline
                Class Name in Arabic &Class Name in English\\
                \hline
                \RL{فسخ نكاح لعوض} & \thead{annulment of marriage for compensation}
\\
                 \hline
                \RL{فسخ نكاح بدون عوض} & annulment of marriage without compensation\\
                \hline
                \RL{فسخ نكاح}&  annulment of marriage\\
                \hline
                \RL{رد دعوة المدعي} & deny Annulment\\
                \hline
                \end{tabular}
                \caption{Judgement Decision Classes in the Annulment of Marriage Cases}
                \label{Classes in annulment of marriage Cases}
                \end{table}

%%%%%%%%%%%%%%%%%%%%%%%%%%%%%%%%%%%%%%%%%
\begin{table}
\begin{adjustbox}{width=15 cm,center}

\begin{tabular}{|c|c|c|}
\hline
\textbf{Type of text} & \textbf{Arabic original text} &\textbf{Translated text}\\
\hline

Plaintiff claim (\RL{نص الدعوى }) & \RL{ادعت المدعية على الغائب مجلس الحكم بأنه كان...} &  \thead{ plaintiff claimed that the defendant \\ (absent from the Governing Council) is..}\\%\hline

Defendant answer(\RL{نص الاجابة})& \RL{وبعرض ذلك على المدعي اجاب قائلا ماذكرت...} & \thead{ by presenting it to the defendant, he replied..}
\\%\hline

Pleading (\RL{نص المرافعة})&  \RL{ما ذكرته المدعية من الزواج والأولاد ثم الطلاق فهذا كله صحيح، ولكن الطلاق...} & \thead{what the wife mentioned about marriage\\ and divorce is true, but the divorce..}\\%\hline

Reasons(\RL{الأسباب})& \RL{ ما صح عن النبي صلى الله عليه وسلم خير غلاما بين أبيه وأمه} & \thead{The Prophet, PBUH, asked a boy to \\ choose between his mother and father..}\\%\hline

Judgment Decision(\RL{ الحكم})& \RL{الحكم بحضانة البنت للأم} & \thead{ mother shall grant custody of her daughter}\\\hline

Plaintiff claim (\RL{نص الدعوى }) & \RL{   ادعى المدعي وكالة قائلا ان الام قد تزوجت  }
& \thead{  plaintiff claimed that the defendant \\ got married}\\  %\hline
Defendant answer(\RL{نص الاجابة}) & \RL{ وبعرض ذلك على المدعي عليها ما ذكره}  & \thead{  by presenting it to the defendant, she replied}\\  %\hline 

Pleading (\RL{ المرافعة})& \RL{موكلي يطلب الحضانة للأسباب التالية }& \thead{My client requested custody for\\ the following reasons..}\\  %\hline
 Reasons(\RL{الأسباب})& \RL{أنت أحق به ما لم تنكحي}& \thead{You are more entitled to him \\ as long as you do not get married.}\\  %\hline
Judgment Decision(\RL{ الحكم})& \RL{الحكم بحضانة البنت للأب} & \thead{ father shall grant custody of his daughter}\\\hline
\end{tabular}
\end{adjustbox}
\caption{example of the multi class classification problem}
\label{Data frame for single label classification}
\end{table}

To increase the model performance, another simulated dataset was collected throughout experts in the law domain, the experts are selected among Master and PhD students, and validated by Arabic speaking law professors. The dataset has a larger number of custody and annulment of marriage cases. As shown in Table~\ref{Simulated dataset} the simulated dataset consist of the plaintiff’s claim, the defendant’s response, pleading (a description of the sessions that took place in court), the judgment, and the evidences (the reasons and legal evidence that was relied upon in judging the case).

%rest of the personal status cases are few

%However, the number of cases in this dataset was low and imbalanced, so we decided to collect more data by writing cases that simulate the cases in the Ministry of Justice dataset.
 %We will explain each one in detail in the following sections.

\begin{table}[h]
\centering
\begin{tabular}{|c|c|}
%\hline
% \multicolumn{2}{|c|}{Simulated dataset}\\ 
 \hline
 \textbf{Source}& Law experts (master, PhD students, Professors)\\
 \hline
 \textbf{Content}&claim, answer, pleading, judgment, evidences\\
 \hline
 \textbf{Types} &\textbf{Number of Cases}\\ 
 \hline
 Custody&55\\
 \hline
 Annulment of marriage&24\\
 \hline
\end{tabular}
\caption{Simulated dataset}
\label{Simulated dataset}
\end{table}

\section{Methodology}\label{sec:Exp}
\subsection{Problem formulation}
 In this study, we have developed two LJP models based on Arabic personal status cases. The first model aims to predict the judgement and reasons (article law) given %the plaintiff claim, defendant answer, and 
 the court pleading of Arabic personal status cases (custody and annulment of marriage). Thus, given a pleading %, claim, and answer 
 of a case $x$ that consist of sequence of words $x= \{x_1,x_2,..,x_k\}$, where $k$ is the length of $x$. The goal is to predict the corresponding judgment result $y_i$ where is one of the possible judgement decisions (the label set in our task) $Y$, $y_i\in Y$. 
 
 The second  model aims to predict the probability of the possible judgment decision and associated reasons given the plaintiff claim and defendant answer. Suppose the claim text $m$ of a case consists of a sequence  of words, $m=\{ m_1,m_2,...m_f\}$, where $f$ is the length of claim text, and the answer text $n$ consists of a sequence of words, $n= \{n_1,n_2,...n_j\}$, where $j$ is the length of the answer. We formulated the this task as a multi label classification problem i which the goal is to predict the probability of each of the corresponding judgment outcomes $y_i \in Y$  where $Y$ is the label set of our task, and the probability of each judgment given $m$ and $n$.

\subsection{Data Preprocessing}
A series of data preprocessing steps were performed on the input data can be clean and normalize the textual data and transform the data into a dense representation that models can analyze. Data preprocessing includes a manual correction of misspelled extracted words, tokenization, data cleaning, including stop words,  unimportant words, dates removal, and text normalization, including diacritical marks removal, were conducted on the dataset.
\begin{enumerate}
%\item Correct wrong words

\item Tokenization

Tokenization is essentially splitting a phrase, sentence, paragraph, or an entire text document into smaller units, such as individual words or terms. Each of these smaller units is called a token. This technique is an important technique because the meaning of the text could easily be interpreted by analyzing the words present in the text.

\item Remove stop words

One of the significant forms of preprocessing is to filter out useless data. In natural language processing, useless words are referred to as "stop words." Eliminating stop words has several advantages, including  shortened indexing structures, faster processing, and improved retrieval effectiveness. The Arabic language has many lexical tokens, which implies stop words can be found in large quantities.

\item Remove dates

The data contains lots of dates which were excluded at this study, such as the date of birth and the date of the session.
\item Remove diacritics\\
Diacritics are marks placed above or below (or sometimes next to) a letter in an Arabic word to indicate a particular pronunciation with regard to accent, tone, or stress, as well as meaning, especially when a homograph exists without the marked letter or letters. In this study, the diacritics were removed to normalize the text, reduce the feature space dimensionality, and ensure there is no difference between a word with diacritics and one without.

\end{enumerate}

\subsection{Text Representation}%\label{Data Representation}
Representing text as a real-valued numerical representation that captures word semantics and similarity between words is an essential step in the natural language process task. In this study, we utilized two widely used word embedding; the first is a discreet text representation in which words are represented by their corresponding indexes to their position in a dictionary from a larger corpus. We, in particular, use a weighted version of the bag of word (BOW) model, namely term frequency inverse document frequency (TF-IDF). TF-IDF normalizes the word frequency by giving frequent words less weight according to the following equation:

\begin{equation}
TF-IDF=TF(w,d)*IDF(w)
\end{equation}

Where $TF(w, d)$ is the frequency of word $w$ in the document $d$ and $IDF(w)$ is defined as follows:
\begin{equation}
IDF(w)=log\frac{N}{IDF(w)}
\end{equation}
Where $N$ is the total number of documents and $df(w)$ is the frequency of documents containing the word $w$. 
The second word representation is distributed word embedding that allows words with similar meanings to have a similar representation. The used distributed text representation is the word2vec model for Arabic language text (Aravec)~\cite{soliman2017aravec}. Word2vec models are a powerful, dense word representation that has been widely used in natural language processing tasks due to their ability to capture word semantic similarities.

\subsection{Model and Hyperparameter}

%\section{Predict the judgment}
We implemented several machine and deep learning models to evaluate the proposed model for predicting judgment and evidences from pleading text. We implemented several machine learning model and deep learning models with TF-IDF and word2vec word representations. Two machine learning are used in our experiment, namely, SVM and LR. While the implemented deep learning are LSTM and BiLSTM.

      \begin{enumerate}
          \item \textbf{SVM Model}\\
We used SVM model due to its simplicity and its efficiency in high dimensional spaces. We applied a grid search to obtain the best hyperparameters C, gamma ('C' for adding a penalty for each misclassified data point, 'gamma' for controlling a single training point's distance of influence), and kernel function that is used for the model. We took the best kernel function from the grid search but used different values for C and gamma to get the best hyper-parameters. %Then fed it to SVC() function and got the result.\\

          \item \textbf{LR Model}\label{LR model1}\\
          We did the same as in SVM, but we used Logistic Regression() with solver= 'sag' that uses the cross-entropy loss and supports the multi-class case. 
          
          \item \textbf{LSTM Model}\label{LSTM Model1}\\
          The architecture of the STM model consists of four layers. The first layer is the input layer with a shape (1200,) and data type 'int32'. The second layer is the embedding layer. The next layer is the LSTM layer with 300 neurons as a parameter, followed by a dense layer with 300 neurons with 'relu' activation functions. The last layer, the dense layer, consists of 4 neurons with a 'softmax' activation function and 'sparse\_categorical\_crossentropy' loss function. For the LSTM model with word2vec word representation, the embedding layer seeded by AraVec word embedding weight (300-dimensional Twitter Skip-gram version 3).
    
    \item \textbf{BILSTM model} \\
    For this model, we used exactly the same hyperparameters as the LSTM model, but we added the Bidirectional wrapper with an LSTM layer (64 neurons) that propagates the input forward and backward through the LSTM layer to learn long-term dependencies from both sides.
      \end{enumerate}

%\section{Predicting the probability of possible outcomes}
To predict the probability of possible outcomes using the claim and answer, we used SVM, LR, LSTM, and BiLSTM with TF-IDF and word2vec text representations. We used the same hyperparameters and methods described previously. In LSTM and BiLSTM, we used four neurons with 'sigmoid' activation function to estimate returns the probability for each class (judgment outcome).

\section{Results and Discussion}\label{sec:Dis}

 \subsection{Evaluation Metrics}
 To evaluate the judgment and evidences prediction, we used accuracy, precision, recall rate, and F1-score.
 \subsection{Results and Analysis}
 We evaluated the performance on two LJP tasks, including predicting the judgment and predicting the probability of possible outcomes. We experimented with different machine learing and deep learning models in order to choose the best models for both tasks. Tables~\ref{Results of experiments for predicting the judgment on Custody and Annulment of Marriage Cases},~\ref{Results of experiments for predicting the evidences in custody and annulment of marriage cases}, and~\ref{Results of experiments for predicting the probability of possible outcomes on custody and annulment of marriage cases} show the experimental results of predicting judgment, law articles, and probability of judgements, respectively. As shown in Table~\ref{Results of experiments for predicting the judgment on Custody and Annulment of Marriage Cases} for predicting the judgment of the custody cases on the dataset that combines data from the Ministry of Justice and simulated data, the SVM model with word2vec representation gave the highest accuracy of 88\%. Deep learning model like LSTM and BiLSTm did not outperform SVM model due to the relatively small size dataset. Table~\ref{Results of experiments for predicting the judgment on Custody and Annulment of Marriage Cases} also shows the results for predicting the judgment of the annulment of marriage cases on the combined dataset. The LR model with TF-IDF data representation give the highest accuracy of 78\%. % We chose the LR model with TF-IDF because it has less over-fitting.
 The results indicate that predicting the judgment in custody cases is higher than the prediction of the annulment of marriage.

\begin{table}[th]
    \centering
    \begin{tabular}{|l|l|l|l|l|l|l|l|l|}
    \hline

             & \multicolumn{4}{c|}{Custody} & \multicolumn{4}{c|}{Annulment of Marriage} \\
      \hhline{~|--------}

        %Custody  & ~ & ~ & ~ &  Annulment of Marriage & ~ & ~ & ~ \\ \hline
        Models & P(\%) & R(\%) & F1(\%) & Acc(\%) & P(\%) & R(\%) & F1(\%) & Acc(\%) \\ \hline
        SVM-TFIDF & 60 & 44.33 & 46.33 & 81 & 62.5 & 63.5 & 62.5 & 78 \\ \hline
        \textbf{SVM-Word2Vec}  & \textbf{88} & \textbf{100 } & \textbf{93} & \textbf{88} & 23.75 & 27.25 & 24.5 & 56 \\ \hline
        LR-TFIDF & 75 & 100 & 86 & 75 & \textbf{62.5 } & \textbf{63.5 } & \textbf{62.5} & \textbf{78 } \\ \hline
        LR-Word2Vec & 86 & 100 & 75 & 75 & 48.75 & 38.5 & 40.75 & 50 \\ \hline
        LSTM-TFIDF & 31 & 33.33 & 29.33 & 87.5 & 24 & 27.75 & 25 & 56.25 \\ \hline
        LSTM-Word2Vec & 31 & 33.33 & 29.33 & 87.5 & 24 & 27.75 & 25 & 56.25 \\ \hline
        BILSTM-TFIDF & 31 & 33.33 & 29.33 & 87.5 & 24 & 27.75 & 25 & 56.25 \\ \hline
        BILSTM-Word2Vec & 75 & 100 & 86 & 75 & 63 & 100 & 77.5 & 62.5 \\ \hline
       % \%AraBert & 31 & 33.33 & 29.33 & 87.5 & 29.5 & 31.75 & 27.5 & 56 \\ \hline
    \end{tabular}
    \caption{ Results of experiments for predicting the judgment on Custody and Annulment of Marriage Cases using machine learning and deep learning models with different word representations}
    \label{Results of experiments for predicting the judgment on Custody and Annulment of Marriage Cases}
\end{table}

\begin{table}[h]
\begin{tabular}{|l|l|l|l|l|l|l|l|l|}
%\begin{tabular}{|lHHH|l|HHHl|}
  % &   &  &  & Custody &  &  &  & Annulment of Marriage     \\ 
\hline
    & \multicolumn{4}{c|}{Custody}   & \multicolumn{4}{c|}{Annulment of Marriage}  \\
         \hhline{~|--------}

Models   & P(\%) & R(\%) & F1(\%) & Acc(\%) & P(\%) & R(\%) & F1(\%) & Acc(\%)    \\ \hline
    SVM-TFIDF & 17.66 & 17 & 10 & 19 & 23.63 & 20.9 & 25 & 34 \\ \hline
        SVM-Word2Vec & 3.66 & 3.66 & 5.55 & 25 & 13 & 12.72 & 10.63 & 25 \\ \hline
        LR-TFIDF & 11.88 & 13.33 & 5 & 12 & 6.81 & 10.81 & 8.27 & 41 \\ \hline
        LR-Word2Vec & 38.88 & 33.33 & 35.22 & 12  & \textbf{6.45 } & \textbf{6.36 } & \textbf{6.18 } & \textbf{50} \\ \hline
        \textbf{LSTM-TFIDF} & \textbf{3.66 } & \textbf{3.66 } & \textbf{5.55 } & \textbf{25 } & 27.57 & 23.05 & 26.06 & 37.5 \\ \hline
        LSTM-Word2Vec & 3.66 & 3.66 & 5.55 & 25 & 27.57 & 23.05 & 26.06 & 37.5 \\ \hline
        \textbf{BILSTM-TFIDF} & \textbf{3.66 } & \textbf{3.66 } & \textbf{5.55 } & \textbf{25 } & 27.57 & 23.05 & 26.06 & 37.5 \\ \hline
        BILSTM-Word2Vec & 40.5 & 34.71 & 36.68 & 12.5 & 12.72 & 15.39 & 15.73 & 30.25 \\ \hline                 
\end{tabular}

\caption{Results of experiments for predicting the evidences (law articles) in custody and annulment of marriage cases using machine learning and deep learning models and text representations}
\label{Results of experiments for predicting the evidences in custody and annulment of marriage cases}
\end{table}

In Table~\ref{Results of experiments for predicting the evidences in custody and annulment of marriage cases}, the results show that the best models for predicting law articles in custody cases are LSTM and BiLSTM with TF-IDF representation with 25\% accuracy. The results of predicting law articles in the annulment of marriage cases were higher, with the best results of 50\% accuracy obtained using LR with word2vec models. These results indicate the challenge of predicting law articles in custody and the annulment of marriage cases.

\begin{table}[h]
\begin{tabular}{|l|l|l|l|l|l|l|l|l|}
%\begin{tabular}{|lHHH|l|HHHl|}
  % &   &  &  & Custody &  &  &  & Annulment of Marriage     \\ 
\hline
    & \multicolumn{4}{c|}{Custody}   & \multicolumn{4}{c|}{Annulment of Marriage}  \\
         \hhline{~|--------}

Models   & P(\%) & R(\%) & F1(\%) & Acc(\%) & P(\%) & R(\%) & F1(\%) & Acc(\%)    \\ \hline
    SVM-TFIDF & 60 & 44.33 & 46.33 & 81 & 94.25 & 91.75 & 98.75 & 68 \\ \hline
        \textbf{SVM-Word2Vec} & \textbf{29.33} & \textbf{33.33} & \textbf{31} & \textbf{88} & 31.66 & 36.33 & 81.33 & 56 \\ \hline
        LR-TFIDF & 25 & 33.33 & 28.66 & 75 & 31.25 & 34.75 & 32.75 & 66 \\ \hline
       \textbf{ LR-Word2Vec} & \textbf{29.33} & \textbf{33.33} & \textbf{31} & \textbf{88} &48.75  & 38.5  & 40.75 & 50 \\ \hline
        LSTM-TFIDF & 29.33 & 33.33 & 31 & 87.5  & 32 & 36.33 & 81.33 & 56.25 \\ \hline
        LSTM-Word2Vec & 29.33 & 33.33 & 31 & 87.5 & 32 & 36.33 & 81.33 & 56.25 \\ \hline
        BILSTM-TFIDF & 29.33  & 33.33 & 31& 87.5  & \textbf{32} & \textbf{36.33} & \textbf{34.33} & \textbf{68.75} \\ \hline
        BILSTM-Word2Vec & 29.33 & 33.33 & 31 & 87.5 & 48.75 & 38.5 & 40.75 & 50 \\ \hline                 
\end{tabular}
\caption{Results of experiments for predicting the probability of possible outcomes on custody and annulment of marriage cases using machine learning and deep learning models}
\label{Results of experiments for predicting the probability of possible outcomes on custody and annulment of marriage cases}
\end{table}

Table~\ref{Results of experiments for predicting the probability of possible outcomes on custody and annulment of marriage cases} shows the results for predicting the probability of judgment outcome in custody cases. SVM and LR with word2vec representation gave the highest accuracy of 88\%. The results also demonstrate that BiLSTM with TF-IDF showed the highest accuracy for predicting the probability of judgment outcome in the annulment of marriage cases with 68\% accuracy. %The LR model with word2vec data representation gives the highest accuracy (50\%).

\section{Conclusion}\label{sec:Con}
Several researchers have developed techniques for predicting the outcomes of cases in the legal profession. However, none of the proposed techniques were implemented in Arabic, and only a few attempts were implemented in English. This project aims to develop an Arabic legal judgment prediction system that utilizes deep learning models, such as LSTM and BiLSTM, and machine learning techniques, such as SVM and LR, to predict the judgment outcome and law articles from Arabic language case scripts, especially personal status cases in Saudi Arabia. %There is also a scarcity of Arabic resources, limiting the field’s advancement. 
Thus, we developed an Arabic LJP dataset from publicly available sample collections combined with the data artificially generated by law experts. The dataset was then analyzed to formulate the LJP task as a multi-class classification problem.
We investigated several machine learning and deep learning models with different types of word representation. The results indicate that for predicting the judgment of the custody cases from pleading text, the SVM model with word2vec data representation achieved the highest accuracy of 88\%. For the annulment of marriage cases, the LR model with TFIDF achieved the best result with 78\% accuracy. For predicting the law article or evidence of custody cases, BiLSTM with TFIDF gave 25\% accuracy, while for the annulment of marriage cases, the LR model with word2vec achieved 50\% accuracy. For predicting the most probable judgment from claim and answer, the highest accuracy of 88\% was obtained using SVM and LR with word2vec representation in custody cases and 68\%  using BiLSTM with TF-IDF in the annulment of marriage cases. 

 %and her students that simulated the available dataset to increase the size of the data overall, then manually corrected and reformulate it to feed it to the model and get accurate results. No personal status law led to the variation in the judgment between judges for the same case. An Arabic legal judgment prediction system was created for the previously mentioned reasons. In conclusion, this project has two major contributions to Arabic LJP: 1) artificially generating an Arabic dataset. 2)	developing an intelligent legal assistant system using transformer-based and ML models to predict judgment results and predict the probability of possible outcomes of a specific case written in Arabic. 
 
%\section{Acknowledgement}

%\bibliographystyle{cas-model2-names}

%\bibliography{ref.bib}

\bibliography{sn-bibliography}% common bib file

\end{document}